\pdfoutput=1

\documentclass[11pt]{article}

\usepackage[final]{acl}

\usepackage{times}
\usepackage{latexsym}

\usepackage[T1]{fontenc}

\usepackage[utf8]{inputenc}

\usepackage{microtype}

\usepackage{inconsolata}

\usepackage{graphicx}

\usepackage{array}
\usepackage{multirow, booktabs, tabularx}
\usepackage{amsmath,amssymb}
\usepackage{bm}
\usepackage{siunitx}

\makeatletter
\def\blfootnote{\xdef\@thefnmark{}\@footnotetext}
\makeatother
\newcolumntype{P}[1]{>{\centering\arraybackslash}p{#1}}
\makeatletter
\def\blfootnote{\xdef\@thefnmark{}\@footnotetext}
\makeatother

\title{Leveraging Cross-Lingual Transfer Learning in \\ Spoken Named Entity Recognition Systems}

\author{
  \textbf{Moncef Benaicha\textsuperscript{1}},
  \textbf{David Thulke\textsuperscript{2}},
  \textbf{M. A. Tu\u{g}tekin Turan\textsuperscript{1}}
\\
\\
  \textsuperscript{1}Fraunhofer Institute for Intelligent Analysis and Information Systems (IAIS), Germany \\
  \textsuperscript{2}Machine Learning and Human Language Technology, RWTH Aachen University, Germany
\\
\\
  \small{
    \textbf{Correspondence:} \href{mailto:moncef.benaicha@rwth-aachen.de}{moncef.benaicha@rwth-aachen.de}
  }
}

\begin{document}
\maketitle
\begin{abstract}
Recent Named Entity Recognition (NER) advancements have significantly enhanced text classification capabilities. This paper focuses on spoken NER, aimed explicitly at spoken document retrieval, an area not widely studied due to the lack of comprehensive datasets for spoken contexts. Additionally, the potential for cross-lingual transfer learning in low-resource situations deserves further investigation. In our study, we applied transfer learning techniques across Dutch, English, and German using both pipeline and End-to-End (E2E) approaches. We employed Wav2Vec2 XLS-R models on custom pseudo-annotated datasets to evaluate the adaptability of cross-lingual systems. Our exploration of different architectural configurations assessed the robustness of these systems in spoken NER. Results showed that the E2E model was superior to the pipeline model, particularly with limited annotation resources. Furthermore, transfer learning from German to Dutch improved performance by 7\% over the standalone Dutch E2E system and 4\% over the Dutch pipeline model. Our findings highlight the effectiveness of cross-lingual transfer in spoken NER and emphasize the need for additional data collection to improve these systems.
\end{abstract}

\section{Introduction}

\blfootnote{This work was carried out while the first author was a research assistant at Fraunhofer IAIS.}

Named Entity Recognition (NER) identifies and classifies named entities within text, including persons, organizations, locations, and other predefined categories \cite{Nadeau07}. While substantial progress has been made in extracting entities from written text, adapting these techniques to spoken content has seen limited research. This is primarily due to the unique challenges associated with spoken text analysis \cite{Tomashenko19}.

Improved spoken NER has significant implications for various practical applications. Accurate entity recognition enhances user interaction in voice assistants by correctly identifying and responding to queries \cite{Jehangir23}. In automatic transcription services, better NER improves the quality of transcriptions by correctly tagging entities, which is crucial for generating accurate and searchable text \cite{Szymanski23}. Spoken dialogue systems in customer service and virtual agents also benefit from enhanced NER by providing more context-aware and accurate responses.

Exploring spoken NER involves challenges due to the unpredictable nature of spoken language. Variabilities in pronunciation, speech disfluencies, and background noise present significant obstacles and, therefore, negatively impact system performance \cite{Porjazovski21}. Additionally, the continuous flow of spoken language, with unclear word boundaries, adds complexity to the task \cite{Chen22}. Despite these difficulties, spoken NER holds significant interest for its potential applications in areas like voice assistants, automatic transcription services, and spoken dialogue systems \cite{Ghannay18,Haghani18,Serdyuk18}.

The introduction of Transformer-driven methodologies has advanced this field significantly. Notably, the End-to-End (E2E) modeling approach directly links speech patterns to transcriptions with embedded entity markers, which show promising results \cite{Mdhaffar22}. These models effectively map temporal dependencies and manage the complexities of various spoken dialects. However, the research primarily focuses on high-resource languages like English, resulting in less effective model performance in data-scarce scenarios.

This paper addresses the issue of linguistic disparity by exploring cross-lingual transfer learning for spoken NER. We focus on using multilingual language representation models to evaluate their effectiveness, especially in data-scarce environments where this term refers to the limited availability of high-quality, manually annotated datasets in comparison to more extensively studied languages like English. Our empirical studies cover three main languages: Dutch, English, German. We specifically examine transfers between languages with different resource levels, highlighting the strength of transfer learning in scenarios from zero to low resources. These languages were chosen for our study due to their varying resource levels and linguistic similarities, which provide a meaningful context for examining cross-lingual transfer learning.

Moreover, we provide a comparative analysis of different methodologies, including both pipeline and E2E approaches. The pipeline framework integrates the functionalities of Automatic Speech Recognition (ASR) systems with subsequent NER models. Initially, the ASR system transcribes spoken content into text, which is then tagged with entities by the NER system. Previous work has explored E2E strategies that simultaneously address ASR and NER tasks \cite{Caubriere20}. This approach aims to refine ASR alongside Natural Language Understanding (NLU) and significantly reduce error propagation commonly caused by ASR limitations \cite{Jannet15}.

A major challenge with E2E models is their need for extensive training datasets. The limited availability of audio-textual datasets with entity annotations emphasizes this issue, as creating large annotated speech datasets is both complex and costly. To address this, \citet{Pasad22}, used a labeling model to generate pseudo-annotations. Inspired by this approach, our paper includes custom pseudo-annotated datasets in Dutch, English, and German, created using the XLM-R\textsubscript{L}-based NER model \cite{goyal2021largerscale}. These pseudo-annotations were not manually corrected, and all evaluations were performed on these pseudo-annotated datasets. Furthermore, no gold-standard annotations were used.

We further examine the impact of various factors such as training data volume, language model choice, and target language, on spoken NER system performance. This paper enhances research in spoken NER by highlighting challenges and potential cross-lingual solutions. Our results advance spoken document retrieval and support the development of more sophisticated and accurate spoken NER systems. Consistent with open research principles, all code, data, and results from this study will be publicly available\footnote{\url{https://github.com/moncefbenaicha/spoken-ner}}. 

To summarize our contributions:
\begin{itemize}
\item We comprehensively compare and analyze pipeline versus E2E strategies for spoken NER in Dutch, English, and German.
\item We investigate transfer learning within both pipeline and E2E strategies for spoken NER.
\item We move from a high-resource language, German, to a resource-scarce language, Dutch, resulting in a notable 10\% improvement in spoken NER performance.
\end{itemize}

\section{Related Work}
\label{sec:literature}

Traditionally, NER from spoken content has utilized a pipeline approach, beginning with an Automatic Speech Recognition (ASR) phase followed by NER on the resulting transcriptions \cite{Jannet17}. While such a system may seem intuitive, it has essential challenges. Specifically, it directly incorporates transcriptions annotated with entities within the ASR system \cite{Cohn19}. By embedding such annotations, there's potential to refine the partial hypotheses, which often get overlooked or dismissed during the decoding process. A novel solution has been integrating specific entity expressions into the lexicon to enhance language model accuracy in recognizing these expressions \cite{Hatmi13}.

In response to these challenges, interest in the E2E approach for spoken NER has increased. This method aims to simultaneously optimize ASR and NER processes, providing a potentially more efficient alternative to the traditional pipeline by leveraging deep neural networks' capabilities to manage long-range sentence dependencies.

\begin{figure}[!b]
\centering
\includegraphics[width=0.88\columnwidth]{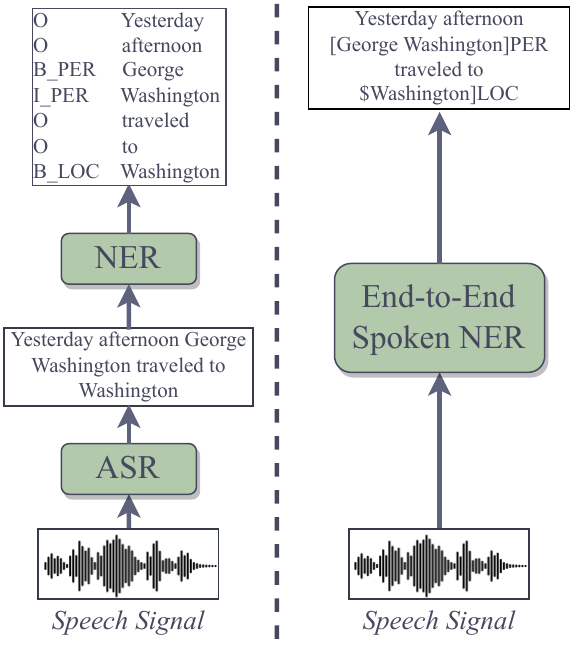}
\caption{The diagram on the left shows a two-stage pipeline, while the right shows the E2E approach.}
\label{fig:block}
\end{figure}

Significant research into the E2E approach for spoken NER includes work with French datasets \cite{Ghannay18}, adopting architectures similar to DeepSpeech \cite{Amodei16}, guided by the Connectionist Temporal Classification (CTC) objective \cite{Graves06}. Building on this, \citet{Yadav20} developed a method tailored for English, introducing specific tokens in the ASR vocabulary to enhance NER tagging. Our research also incorporates this by incorporating unique symbols ('\{', '[', '\$', ']') in transcripts to assist in identifying entities such as organizations, persons, and locations, as depicted in Figure~\ref{fig:block}. Building on these foundational studies, more recent research \cite{Shon22,Pasad22} has successfully employed this methodology in conjunction with the Wav2Vec2 model.

Consequently, the E2E strategy aims to align speech utterances with annotated transcriptions perfectly, facilitating direct entity extraction from spoken content. Empirical studies using French and English datasets have demonstrated the effectiveness of the E2E strategy, often surpassing traditional pipeline approaches, particularly against Long Short-term Memory (LSTM) based models, which no longer meet state-of-the-art standards \cite{Vajjala}.

\section{Methodology}
\label{sec:methodology}

\subsection{Baseline Models}


This paper introduces two distinct baseline systems designed for English, German, and Dutch. The first system adopts a conventional pipeline approach where ASR and NER models are trained separately. After training, these models are integrated during the inference phase to produce results. In contrast, the second system employs an E2E methodology involving more intricate processes. For the E2E system, the ASR model is fine-tuned using the robust pre-trained Wav2Vec2-XLS-R-300M model \cite{babu2021xlsr}. During this fine-tuning, each language is treated separately, utilizing CTC-loss as the main objective function. Concurrently, the NER component of the pipeline is enhanced through modifications to the XLM-R\textsubscript{L} language representation model. This enhancement includes the addition of a linear layer specifically designed to handle lower-cased tokens from the CoNLL 2002 and 2003 datasets \cite{sang2002introduction, sang2003introduction}.


To enhance the ASR model, we integrate a 4-gram language model trained on both the training and development datasets for each language. In the E2E system, we utilize the capabilities of the Wav2Vec2-XLS-R-300M model once more. For this iteration, we fine-tune using a specially augmented corpus. This corpus includes special tokens that explicitly indicate the start of an entity such as a person (PER), an organization (ORG), or a location (LOC) and mark the end of an entity mention, as shown in Figure~\ref{fig:block}.

\begin{table*}[!t]
\centering
\renewcommand{\arraystretch}{1.1}
\begin{tabular}{l *{9}{c}}
\toprule
& \multicolumn{3}{c}{Training Set} & \multicolumn{3}{c}{Validation Set} & \multicolumn{3}{c}{Test Set} \\
\cmidrule(lr){2-4} \cmidrule(lr){5-7} \cmidrule(lr){8-10}
& \textbf{EN} & \textbf{DE} & \textbf{NL} & \textbf{EN} & \textbf{DE} & \textbf{NL} & \textbf{EN} & \textbf{DE} & \textbf{NL} \\
\midrule
\#Sentences       & 527K & 526K & 42K & 5K & 5K & 5K & 5K & 5K & 5K \\
\#Tokens          & 5.4M & 5M & 0.5M & 46K & 47K & 46K & 47K & 45.6K & 45K \\
\#Tokens as LOC   & 211K & 156K & 588 & 1.2K & 1.5K & 241 & 1.3K & 1.3K & 314 \\
\#Tokens as ORG   & 177K & 83K & 95 & 1K & 729 & 66 & 1.1K & 621 & 88 \\
\#Tokens as PER   & 216K & 144K & 104 & 1.3K & 1.5K & 123 & 1.4K & 1.2K & 155 \\
\#Tokens as O     & 4.8M & 4.6M & 408K & 43K & 43K & 45K & 43K & 42.4K & 44.5K \\
Total Hours       & 840 & 839 & 54 & 8 & 8.5 & 6.5 & 8.5 & 8.5 & 7 \\
\bottomrule
\end{tabular}
\caption{The statistics of pseudo-annotated data across English, German, and Dutch splits.}
\label{tab:spoken-ner-data}
\end{table*}

\subsection{Transfer Learning Models}


In the pipeline approach, we leverage a dual-component architecture. Specifically, we replace the native NER model with a cross-lingual variant that is more universally applicable. In our transfer learning experiments, German serves as the source language, while English and Dutch are designated as target languages. The first step involves applying the pre-trained German NER model directly to the English and Dutch transcripts without any language-specific modifications. This initial approach is followed by a phase that combines transfer learning with model fine-tuning. During this phase, the German NER model forms the foundation. Fine-tuning is then performed using a subset ($k$) of the target language's training set, corresponding to either 10\% or 20\% of the original dataset. The model's performance is assessed using the test set of the target language to evaluate its effectiveness after fine-tuning.


In contrast, the E2E model adopts a more comprehensive transfer learning approach. Our experiments begin with a zero-shot transfer learning phase, where the capabilities of the German E2E spoken NER system are extended to English and Dutch. Following this initial phase, we apply an extensive fine-tuning using a portion of the target language’s training dataset, typically 20\% or 40\% of the total. After completing this fine-tuning, the model is assessed against the test set of the target language. This evaluation yields critical insights into the effectiveness of the fine-tuning process.

\section{Corpus Overview}
\label{sec:data}

For our experiments, we utilized data from Common Voice\footnote{\href{https://commonvoice.mozilla.org/en/datasets}{https://commonvoice.mozilla.org/en/datasets}}, specifically selecting the validated corpus for each language. This corpus comprises various types of oral data, including read speech from diverse demographics. During preprocessing, we removed duplicate entries, retained essential punctuation, and converted non-Latin characters to Latin script, standardizing everything to lowercase. The processed data were then used with the XLM-R\textsubscript{L}-based NER model, trained on the language-specific CoNLL dataset \cite{sang2002introduction, sang2003introduction}, to generate pseudo-annotations. Note that, in the English corpus, all punctuation except apostrophes was removed during the transcription process.

We encountered non-Latin scripts such as Cyrillic or Brahmic during preprocessing due to the inclusion of multilingual text data in the Common Voice corpus. These scripts were converted to Latin to maintain consistency across datasets. Detailed statistics of the training, development, and testing sets, including the number of sentences, tokens, and total hours, are presented in Table~\ref{tab:spoken-ner-data}. This table illustrates the comprehensive scope and linguistic variability of the datasets used.

Additionally, Table~\ref{tab:entities_overlap} highlights the extent of entity overlaps within and between the languages studied. These were computed by comparing the exact match of entity spans across different languages. While overlaps within each language's training sets are complete, the inter-language overlaps vary significantly. For instance, the overlap between English and German is 36.5\%, contrasting sharply with the minimal 0.1\% overlap between English and Dutch. These statistics underscore the challenges and considerations in developing multilingual NER systems that can effectively transfer learning across languages.


\section{Evaluation Metrics}
\label{sec:metrics}

\begin{table}[!b]
\centering
\renewcommand{\arraystretch}{1.1}
\begin{tabular}{>{\raggedright\arraybackslash}p{1cm} >{\centering\arraybackslash}p{1cm} >{\centering\arraybackslash}p{1cm} >{\centering\arraybackslash}p{1cm} >{\centering\arraybackslash}p{1cm}}
\toprule
\multicolumn{2}{c}{}  & \multicolumn{3}{c}{Train} \\
\cmidrule{3-5}
                       &       & \textbf{EN}       & \textbf{DE}       & \textbf{NL}       \\
\midrule
\multirow{3}{*}{Train} & \textbf{EN}    & 100.0    & 24.7     & 0.1      \\
                       & \textbf{DE}    & 36.5     & 100.0    & 0.2      \\
                       & \textbf{NL}    & 52.4     & 58.0     & 100.0    \\
\midrule
\multirow{3}{*}{Test}  & \textbf{EN}    & 79.6     & 64.0     & 1.7      \\
                       & \textbf{DE}    & 58.4     & 84.3     & 2.1      \\
                       & \textbf{NL}    & 64.3     & 68.1     & 27.0     \\
\bottomrule
\end{tabular}
\caption{The percentages of entity overlaps across different languages.}
\label{tab:entities_overlap}
\end{table}




We employ various metrics to assess the performance of our spoken NER systems. The conventional Word Error Rate (WER) serves as the primary metric for evaluating the accuracy of ASR models. In addition, we utilize the Entity Error Rate (EER), which evaluates the specific accuracy of our spoken NER systems in the context of entity transcription. Unlike broader measures, EER focuses exclusively on the accuracy with which entities such as names of people, locations, and organizations are transcribed:

\begin{equation}
    EER = \frac{N_{\text{{Incorrectly Transcribed Entities}}}}{N_{\text{{Total Entities}}}}
\end{equation}

Here, \(N_{\text{{Incorrectly Transcribed Entities}}}\) denotes the number of entities that the system has transcribed incorrectly, while \(N_{\text{{Total Entities}}}\) indicates the total number of entities in the dataset. The EER measures the proportion of entities that were inaccurately transcribed, providing a direct indicator of the system's transcription accuracy.

Furthermore, to evaluate the effectiveness of the NER system, we calculate the micro-average F1-score, which is a harmonized measure of precision and recall, similar to the approach in \cite{Pasad22}:

\begin{equation}
    F_1 = \frac{TP}{TP+\frac{1}{2}(FP+FN)}
\end{equation}
where true positives (TP) are the correctly identified entities, false positives (FP) are the incorrectly identified entities, and false negatives (FN) are the entities that were not identified. These values are computed based on the accuracy of entity transcription, their types, and the positions of these entities within the transcript. This score provides a balanced measure of the system's overall accuracy in recognizing and classifying entities.

\section{Experiments and Results}
\label{sec:experiments}


In our comprehensive baseline experiments for the pipeline model, we utilized prominent open-source pre-trained models to enhance performance across various tasks. For ASR, we selected the Wav2Vec2-XLS-R-300M pre-trained model\footnote{\href{https://hf.co/facebook/wav2vec2-xls-r-300m}{https://hf.co/facebook/wav2vec2-xls-r-300m}}, which is noted for its efficiency. Simultaneously, for NER, we employed the robust XLM-R\textsubscript{L} pre-trained language model\footnote{\href{https://hf.co/xlm-roberta-large}{https://hf.co/xlm-roberta-large}}, well-known for its effectiveness in multilingual processing.


For fine-tuning the ASR model, we chose the AdamW optimizer, following recommendations by \citet{Loshchilov}, with $betas=(0.9, 0.999)$ and a precision parameter of $eps=10^{-8}$. The learning rate was carefully managed through a schedule that includes a warm-up phase covering one-third of the total training steps. During this phase, the learning rate gradually increases from a baseline to a peak of $10^{-4}$. After this warm-up period, the learning rate linearly decreases through the remaining training sessions. It is important to highlight that we only froze the feature encoder of the Wav2Vec2 architecture and made no other structural changes to the model.


Regarding the NER model, we mostly maintained the same optimizer settings as those used for the ASR model, with a slight adjustment to the maximum learning rate of the scheduler, setting it to $2\times 10^{-5}$. Finally, for the E2E spoken NER model, we chose to maintain consistency by adopting the same set of hyper-parameters previously applied in the ASR component of pipeline framework.

\subsection{Baseline Results}

Table~\ref{tab:baseline_results} presents the baseline performances across varying configurations and languages. It's evident from the results that the EER and the F1-scores consistently display a negative correlation across all three languages. This observed trend aligns with our expectations. Specifically, even if the ASR system exhibits any discrepancies in transcribing an entity accurately, it adversely affects the F1-score. This impact remains regardless of whether the entity boundaries and categories are delineated correctly.

Focusing on the German language, the E2E baseline results stand out. Out of a total of 5000 test utterances for German, 2928 have been transcribed with precision, translating to an accuracy rate of approximately 58.6\% of the test set. Impressively, within this accurately transcribed set, a staggering 94\% of utterances have been labeled correctly, a testament to the model's efficacy. When we pivot our attention to the subset of utterances that weren't transcribed with utmost accuracy, it's important to note that the model, despite the transcription issues, managed to label entities correctly in about 51.3\% of these instances. Cumulatively, this implies that the E2E model adeptly labeled entities with accuracy in nearly 76\% of the entire set of test utterances for German.

It's also worthwhile to note the relative performances of English and Dutch. For instance, when observing the pipeline approach with both ASR and NER components active, English has a WER of 16.7\% and an F1-score of 40.7\%, whereas Dutch, under the same conditions, recorded a WER of 9.3\% and an F1-score of 40.0\%. Such statistics offer nuanced insights into the distinct challenges and variances inherent to each language, emphasizing the importance of tailored strategies for each linguistic domain.

\begin{table}[!t]
\centering
\setlength{\tabcolsep}{5pt} 
\renewcommand{\arraystretch}{1.1} 
\begin{tabularx}{\columnwidth}{l *{6}{>{\centering\arraybackslash}X}} 
\toprule
\textbf{Lang} & \textbf{E2E} & \textbf{ASR} & \textbf{NER} & \textbf{WER} & \textbf{EER} & \textbf{F1}  \\ 
\midrule
\multirow{3}{*}{\textbf{EN}} & no  & no  & yes & N/A  & N/A  & 80.3 \\ 
                             & no  & yes & yes & 16.7 & 48.0 & 40.7 \\ 
                             & yes & no  & no  & 16.5 & 46.0 & 41.8 \\ 
\midrule
\multirow{3}{*}{\textbf{DE}} & no  & no  & yes & N/A  & N/A  & 87.2 \\ 
                             & no  & yes & yes & 9.4  & 29.0 & 61.1 \\ 
                             & yes & no  & no  & 9.1  & 27.0 & 61.6 \\ 
\midrule
\multirow{3}{*}{\textbf{NL}} & no  & no  & yes & N/A  & N/A  & 85.6 \\ 
                             & no  & yes & yes & 9.3  & 49.0 & 40.0 \\ 
                             & yes & no  & no  & 9.2  & 47.0 & 37.4 \\ 
\bottomrule
\end{tabularx}
\caption{Performances of baseline models across different configurations and languages.}
\label{tab:baseline_results}
\end{table}

A comparable trend is also observed in the pipeline system, with the E2E model exhibiting a marginally superior performance. One of the E2E model's significant strengths lies in its ability to tag entities accurately, even in the presence of transcription errors. This proficiency translates to a performance uptick of around 2\% when pitted against the traditional pipeline system. Exploring the performances across different languages, our analyses of English and Dutch mirror the patterns we uncovered for German. However, the Dutch distinguishes itself in one noteworthy aspect: an impressive accuracy rate of up to 99\% in entity identification when the transcriptions are accurate. Across the board, the E2E spoken NER system demonstrates a propensity to fine-tune the WER, EER, and by extension, the F1-score. The sole deviation from this pattern is seen in Dutch. This anomaly can be attributed to the relatively limited pool of training data available for Dutch, especially when benchmarked against English and German.

\subsection{Transfer Learning Results}


Tables \ref{tab:transfer_learning_results_de_en} and \ref{tab:transfer_learning_results_de_nl} present the results of transfer learning from German to English and German to Dutch, respectively. These results highlight patterns consistent with our baseline pipeline experiments.



In the German-to-English transition detailed in Table~\ref{tab:transfer_learning_results_de_en}, the performance of E2E model in the zero-shot transfer learning scenario illustrates the adaptability of this approach. However, for the transition from German to Dutch, the situation is markedly different. Zero-shot transfer learning outcomes for German to Dutch align closely with baseline performances, with WER, EER, and F1-scores remaining relatively stable across the scenarios. Interestingly, the incorporation of 40\% of the Dutch training data leads to a noticeable improvement in performance, particularly a roughly 2\% increase in F1-scores as shown in Table~\ref{tab:transfer_learning_results_de_nl}.

Focusing again on the E2E model, it is apparent from the results that the German-to-Dutch transfer yields better performance metrics compared to the German-to-English transfer. An examination of Table~\ref{tab:entities_overlap} provides a potential explanation for this difference, indicating a more substantial overlap in entities between German and Dutch than between German and English.

\begin{table}[!b]
    \centering
    \setlength{\tabcolsep}{3pt} 
    \renewcommand{\arraystretch}{1.1} 
    \begin{tabular}{@{}lccccc@{}} 
        \toprule
        \multirow{2}{*}{\textbf{System}} & \multicolumn{2}{c}{\textbf{Transfer Learning}} & \multirow{2}{*}{\textbf{WER}} & \multirow{2}{*}{\textbf{EER}} & \multirow{2}{*}{\textbf{F1}}  \\
        \cmidrule(lr){2-3}
        & \footnotesize{\textbf{Source $\rightarrow$ Target}} & \bm{$k$} & & & \\
        \midrule
        \multirow{4}{*}{Pipeline} & N/A & N/A & 16.7 & 48.0 & 40.7 \\
        \cmidrule(lr){2-6}
        & \multirow{3}{*}{DE $\rightarrow$ EN} & 0\% & 16.7 & 48.0 & 38.5 \\
        & & 20\% & 16.7 & 48.0 & 39.6 \\
        & & 40\% & 16.7 & 48.0 & 40.0 \\
        \midrule
        \multirow{4}{*}{E2E} & N/A & N/A & 16.5 & 46.0 & 41.8 \\
        \cmidrule(lr){2-6}
        & \multirow{3}{*}{DE $\rightarrow$ EN} & 0\% & 52.5 & 66.0 & 20.8 \\
        & & 20\% & 21.8 & 53.0 & 35.8 \\
        & & 40\% & 22.9 & 54.0 & 35.8 \\
        \bottomrule
    \end{tabular}
    \caption{Performance of the pipeline and E2E models with German-to-English transfer learning, measured across various metrics.}
    \label{tab:transfer_learning_results_de_en}
\end{table} 



A closer look at Table~\ref{tab:transfer_learning_results_de_nl} reveals that fine-tuning the German E2E system with 40\% of the Dutch training data significantly enhances the system’s effectiveness in recognizing Dutch entities. This fine-tuning results in a performance increase of approximately 7\% compared to the standalone Dutch E2E system and a 4\% improvement over the Dutch pipeline system. The gains are particularly notable in the F1 scores within the PER and LOC entity categories, where there is an impressive 10\% increase compared to the baseline Dutch E2E system. These findings underscore the efficacy of targeted training data in boosting system performance and highlight the benefits of cross-lingual transfer learning in multilingual NER systems.

\begin{table}[!t]
    \centering
    \setlength{\tabcolsep}{3pt} 
    \renewcommand{\arraystretch}{1.1} 
    \begin{tabular}{@{}lccccc@{}} 
        \toprule
        \multirow{2}{*}{\textbf{System}} & \multicolumn{2}{c}{\textbf{Transfer Learning}} & \multirow{2}{*}{\textbf{WER}} & \multirow{2}{*}{\textbf{EER}} & \multirow{2}{*}{\textbf{F1}}  \\
        \cmidrule(lr){2-3}
        & \footnotesize{\textbf{Source $\rightarrow$ Target}} & \bm{$k$} & & & \\
        \midrule
        \multirow{4}{*}{Pipeline} & N/A & N/A & 9.3 & 49.0 & 40.0 \\
        \cmidrule(lr){2-6}
        & \multirow{3}{*}{DE $\rightarrow$ NL} & 0\% & 9.3 & 49.0 & 40.2 \\
        & & 20\% & 9.3 & 49.0 & 40.3 \\
        & & 40\% & 9.3 & 49.0 & 42.0 \\
        \midrule
        \multirow{4}{*}{E2E} & N/A & N/A & 9.2 & 47.0 & 37.4 \\
        \cmidrule(lr){2-6}
        & \multirow{3}{*}{DE $\rightarrow$ NL} & 0\% & 78.0 & 67.0 & 24.0 \\
        & & 20\% & 11.8 & 42.0 & 44.4 \\
        & & 40\% & 10.7 & 40.0 & 44.3 \\
        \bottomrule
    \end{tabular}
    \caption{Performance of the pipeline and E2E models with German-to-Dutch transfer learning, measured across various metrics.}
    \label{tab:transfer_learning_results_de_nl}
\end{table}

\section{Conclusions}
\label{sec:conc}

In this paper, we explore spoken NER with a focus on cross-lingual transfer learning, employing both pipeline and E2E methodologies. Our findings indicate that the E2E approach to spoken NER generally outperforms the pipeline method in terms of both diverse evaluation metrics and overall parameter efficiency. Nevertheless, the pipeline approach retains its practical utility due to its flexibility in integrating various ASR and NER components.

Our investigations show that deploying a German NER model without fine-tuning in a Dutch or English context within the pipeline still allows the E2E spoken NER to achieve comparable or superior results to an NER model trained specifically for the pipeline's target language. This highlights the effectiveness of transfer learning in E2E spoken NER systems, which often surpass the performance of traditional pipeline systems. A key insight from our study is the robustness of the E2E model in tagging entities correctly, even when faced with transcription errors, slightly outperforming the pipeline approach.

Looking ahead, several promising directions for further research have emerged. One potential area involves refining the objective function of the ASR model to enhance focus on specific tokens within transcriptions that are of greater relevance to NER tasks. Another promising direction is the investigation of spoken NER within a multilingual framework that can accommodate a wide range of languages and dialects, potentially making significant advancements in the field. Additionally, creating and using human-annotated datasets, with consistent entity annotations across various languages, are crucial. We develop human-annotated datasets where such datasets would provide a solid foundation for evaluating spoken NER systems.

\blfootnote{This work was supported by the European Union's Horizon 2020 Research and Innovation Program under Grant Agreement No. 957017, \url{https://selma-project.eu}.}


\bibliography{custom}

\end{document}